%% file: main.tex
\documentclass[runningheads]{llncs}
\usepackage[T1]{fontenc}
\usepackage{graphicx}
\usepackage{color}
\usepackage{amssymb}
\usepackage[linesnumbered,ruled,vlined]{algorithm2e}

\SetCommentSty{mycommfont}
\SetKwInput{KwInput}{Input}                
\SetKwInput{KwOutput}{Output}              
\SetKwInput{KwFunctions}{Functions}         
\SetKwInput{KwTrainables}{Trainable Parameters}
\SetKwInput{KwStopgrad}{Stop Gradient} 
\SetKwInput{KwHyperparameters}{Hyperparameters}    
\SetKwInput{KwSamples}{Samples}   

\usepackage{tablefootnote}
\usepackage{booktabs, multirow} 
\usepackage{soul}
\usepackage[table]{xcolor} 
\usepackage{changepage,threeparttable} 
\usepackage[colorlinks=true,urlcolor=blue,linkcolor=blue,citecolor=blue]{hyperref}
\usepackage{array}
\usepackage{xcolor}
\usepackage{makecell}
\usepackage{epigraph}
\def\figurename{Fig.}
\def\sectionname{Sec.}
\def\tablename{Table}
\newcolumntype{P}[1]{>{\centering\arraybackslash}p{#1}}
\definecolor{maroon}{RGB}{58, 183, 149}
\definecolor{iblue}{RGB}{33, 53, 236 }
\definecolor{tablered}{cmyk}{0,0.87,0.68,0.32}
\newcommand{\etal}{\mbox{\textit{et al.}}}
\newcommand{\ie}{\mbox{\textit{i.e.,\ }}}
\newcommand{\eg}{\mbox{\textit{e.g.,\ }}}

\begin{document}
\title{Foundation Ark: Accruing and Reusing Knowledge for Superior and Robust Performance}

\titlerunning{Foundation Ark: Accruing and Reusing Knowledge}
%
\author{DongAo Ma\inst{1} \and 
 Jiaxuan Pang\inst{1} \and
Michael B. Gotway\inst{2} \and
Jianming Liang\inst{1}} 
\authorrunning{D. Ma et al. }
\institute{Arizona State University, Tempe, AZ 85281, USA 
\email{\{dongaoma,jpang12,jianming.liang\}@asu.edu} \and
 Mayo Clinic, Scottsdale, AZ 85259, USA\\
\email{Gotway.Michael@mayo.edu}}
\maketitle              
\begin{abstract}
Deep learning nowadays offers expert-level and sometimes even super-expert-level performance, but achieving such performance demands massive annotated data for training (\eg Google's \textit{proprietary} CXR Foundation Model (CXR-FM) was trained on 821,544 \textit{labeled} and mostly \textit{private} chest X-rays (CXRs)). \textit{Numerous} datasets are \textit{publicly} available in medical imaging but individually \textit{small} and \textit{heterogeneous} in expert labels. We envision a powerful and robust foundation model that can be trained by aggregating numerous small public datasets. To realize this vision, we have developed {\bf Ark}, a framework that {\bf a}ccrues and {\bf r}euses {\bf k}nowledge from \textit{heterogeneous} expert annotations in various datasets. As a proof of concept, we have trained two Ark models on 335,484 and 704,363 CXRs, respectively, by merging several datasets including ChestX-ray14, CheXpert, MIMIC-II, and VinDr-CXR, evaluated them on a wide range of imaging tasks covering both classification and segmentation via fine-tuning, linear-probing, and gender-bias analysis, and demonstrated our Ark's superior and robust performance over the state-of-the-art (SOTA) fully/self-supervised baselines and Google's proprietary CXR-FM. This enhanced performance is attributed to our simple yet powerful observation that aggregating numerous public datasets diversifies patient populations and accrues knowledge from diverse experts, yielding unprecedented performance yet saving annotation cost. With all codes and pretrained models released at \url{GitHub.com/JLiangLab/Ark}, we hope that Ark exerts an important impact on open science, as accruing and reusing knowledge from expert annotations in public datasets can potentially surpass the performance of proprietary models trained on unusually large data, inspiring many more researchers worldwide to share codes and datasets to build open foundation models, accelerate open science, and democratize deep learning for medical imaging.

\end{abstract}
\section{Introduction}
Deep learning nowadays offers expert-level and sometimes even super-expert-level performance, deepening and widening its applications in medical imaging and resulting in numerous public datasets for research, competitions, and challenges. 
These datasets are generally small as annotating medical images is challenging, but achieving superior performance by deep learning demands massive annotated data for training. For example, Google's \textit{proprietary} CXR Foundation Model (CXR-FM) was trained on 821,544 \textit{labeled} and mostly \textit{private} CXRs~\cite{sellergren2022simplified}. We hypothesize that powerful and robust \textit{open} foundation models can be trained by aggregating numerous small \textit{public} datasets.
To test this hypothesis, we have chosen CXRs because they are one of the most frequently used modalities, and our research community has accumulated copious CXRs (see \tablename~\ref{tab:datasets}). However, annotations associated with these public datasets are inconsistent in disease coverage. Even when addressing the same clinical issue, datasets created at different institutions tend to be annotated differently. For example, VinDr-CXR~\cite{nguyen2022vindr} is associated with global (image-level) and local (boxed-lesions) labels, while MIMIC-CXR~\cite{irvin2019chexpert} has no expert labels \textit{per se} but comes with radiology reports. ChestX-ray14~\cite{wang2017nihchestx} and CheXpert~\cite{irvin2019chexpert} both cover 14 conditions at the image level, and their 14 conditions have overlaps but are not exactly the same. Therefore, this paper seeks to address a critical need: \textit{How to utilize a large number of \ul{publicly-available} images from different sources and their readily-accessible but \ul{heterogeneous} expert annotations to pretrain generic source (foundation) models that are more robust and transferable to application-specific target tasks}.

To address this need, we have developed a framework, called  {\bf Ark} for its ability of {\bf a}ccruing and {\bf r}eusing {\bf k}nowledge embedded in heterogeneous expert annotations with numerous datasets, as illustrated in \figurename~\ref{fig:ark}. We refer to the pretrained models with Ark as Foundation Ark or simply as Ark for short. To demonstrate Ark's capability, we have trained two models: Ark-5 on Datasets 1-5 and Ark-6 on Datasets 1-6 (\tablename~\ref{tab:datasets}), evaluated them on a wide range of 10 tasks via fine-tuning and on 6 tasks via linear probing, and demonstrated our Ark models outperform the SOTA fully/self-supervised baselines (\tablename~\ref{tab:finetune}) and Google CXR-FM\footnote[1]{\scriptsize\url{GitHub.com/Google-Health/imaging-research/tree/master/cxr-foundation}}
(\figurename~\ref{fig:linear}). Ark also exhibits superior robustness over CXR-FM in mitigating underdiagnosis and reducing gender-related biases, with lower false-negative rates and greater robustness to imbalanced data (\figurename~\ref{fig:bias}).

This performance enhancement is attributed to a simple yet powerful observation that aggregating numerous public datasets costs nearly nothing but enlarges data size, diversifies patient populations, and accrues expert knowledge from a large number of sources worldwide; thereby offering unprecedented performance yet reducing annotation cost. More important, Ark is fundamentally \textit{different} from self-supervised learning (SSL) and federated learning (FL) in concept. SSL can naturally handle images from different sources, but their associated expert annotations are left out of pretraining~\cite{Ma2022Benchmarking}. Clearly, every bit of expert annotation counts, conveying valuable knowledge. FL can utilize data with annotations from different sources, typically involving homogeneous labels, but it mainly concerns data privacy~\cite{mcmahan2017communication}. By contrast, Ark focuses on heterogeneous expert annotations with public data with no concern for data privacy and employs centralized training, which usually offers better performance with the same amount of data and annotation than distributed training as in FL.

Through this work, we have made the following contributions:
    ({\bf 1}) An idea that aggregates public datasets to enlarge and diversify training data;
    ({\bf 2}) A student-teacher model with multi-task heads via cyclic pretraining that accrues expert knowledge from existing heterogeneous annotations to achieve superior and robust performance yet reduce annotation cost;
    ({\bf 3}) Comprehensive experiments that evaluate our Ark via fine-tuning, linear-probing, and few-shot learning on a variety of target tasks, demonstrating Ark's better generalizability and transferability in comparison with SOTA methods and Google CXR-FM;
    and ({\bf 4}) Empirical analyses for a critical yet often overlooked aspect of medical imaging models---robustness to underdiagnosis and gender imbalance, highlighting Ark significantly enhances reliability and safety in clinical decision-making.

\begin{table}[t]
\centering
\caption{Publicly available datasets are generally small and heterogeneously annotated. Our Ark (\figurename~\ref{fig:ark}) aims to aggregate numerous datasets with heterogeneous annotations to diversify patient population, accrue knowledge from diverse experts, and meet the demand by deep learning for massive annotated training data, offering superior and robust performance (\tablename~\ref{tab:finetune}, \figurename~\ref{fig:linear} and \figurename~\ref{fig:bias}) yet reducing annotation cost.}
\label{tab:datasets}
\scriptsize
\begin{tabular}{llllll}\toprule
Abbrev. &Dataset &Task &Usage*&(Pre)train/val/test \\\midrule
1.CXPT &CheXpert~\cite{irvin2019chexpert} &classify 14 thoracic diagnoses  &P|F|L|B &223414/-/234\\
2.NIHC &NIH ChestX-ray14~\cite{wang2017nihchestx} &classify 14 thoracic diseases &P|F|L|B &75312/11212/25596\\
3.RSNA &RSNA Pneumonia~\cite{kaggle2018RSNAPneumonia} &classify lung opacity, abnormality &P|F|L &21295/2680/2709\\
4.VINC &VinDr-CXR~\cite{nguyen2022vindr} &classify 6 thoracic diagnoses &P|F|L &15000/-/3000\\
5.NIHS &NIH Shenzhen CXR~\cite{jaeger2014shenzhenmontgomery} &classify tuberculosis &P|F|L &463/65/134\\
6.MMIC &MIMIC-II~\cite{johnson2019mimic} &classify 14 thoracic diagnoses$^{\dagger}$ & P &368879/2992/5159\\
7.NIHM &NIH Montgomery~\cite{jaeger2014shenzhenmontgomery} &segment lungs &F &92/15/31\\
8.JSRT &JSRT~\cite{shiraishi2000jsrt} &segment lungs, heart, clavicles &F &173/25/49\\
9.VINR &VinDr-RibCXR~\cite{nguyen2021vindrrib} &segment 20 ribs &F &196/-/49\\
10.SIIM &SIIM-ACR PTX~\cite{siim-acr-pneumothorax-segmentation} &classify pneumothorax$^{\ddagger}$ &L &10675/-/1372\\
\bottomrule
\end{tabular}
\begin{tablenotes} 
\item * The usage of each dataset in our experiments is denoted with P for pretraining, F for fine-tuning, L for linear probing, and B for bias study.
\item $^{\dagger}$ The labels of CXRs in MIMIC-II are derived from their corresponding radiology reports using NegBio~\cite{peng2018negbio} and CheXpert~\cite{irvin2019chexpert}.
\item $^{\ddagger}$ SIIM-ACR, originally for pneumothorax segmentation, is converted into a classification task for linear probing, as CXR-FM cannot be evaluated for segmentation using its only released API. 
\end{tablenotes}
\end{table}

\section{Accruing and Reusing Knowledge}
\label{sec:ark}
Our Ark aims to learn superior and robust visual representations from large-scale {\em aggregated} medical images by accruing and reusing the expert knowledge embedded in all available {\em heterogeneous} labels. The following details our Ark.

\smallskip
\noindent\textbf{Accruing knowledge into the student via cyclic pretraining. }
A significant challenge with training a single model using numerous datasets created for different tasks is label inconsistency (\ie heterogeneity) (see~\tablename~3 in Appendix). Manually consolidating heterogeneous labels from different datasets would be a hassle. To circumvent this issue, for each task, we introduce a specific classifier, called task head, to learn from its annotation and encode the knowledge into the model. A task head can be easily plugged into Ark, making Ark scalable to additional tasks. With multi-task heads, Ark can learn from multiple tasks concurrently or cyclically. In concurrent pretraining, a mini-batch is formed by randomly sampling an equal number of images from each dataset, and the loss for each image is computed based on its associated dataset id and labels. This idea is intuitive, but the model hardly converges; we suspect that the loss summation over all task heads simultaneously weakens gradients for back-propagation, causing confusion in weight updating. We opt for cyclic pretraining by iterating through all datasets sequentially in each round to accrue expert knowledge from all available annotations, a strategy that, we have found, stabilizes Ark's pretraining and accelerates its convergence.

\smallskip
\noindent\textbf{Accruing knowledge into the teacher via epoch-wise EMA. }
To further summarize the accrued knowledge and accumulate the learning experiences in the historical dimension, we introduce into Ark a teacher model that shares the same architecture with the student. The teacher is updated using exponential moving average (EMA)~\cite{tarvainen2017mean} based on the student's {\em one epoch of learning} at the end of each task. Eventually, the expert knowledge embedded in all labels and all historical learning experiences are accrued in the teacher model for further reuse in the cyclic pretraining and for future application-specific target tasks.

\begin{figure}[t]
\includegraphics[width=\textwidth]{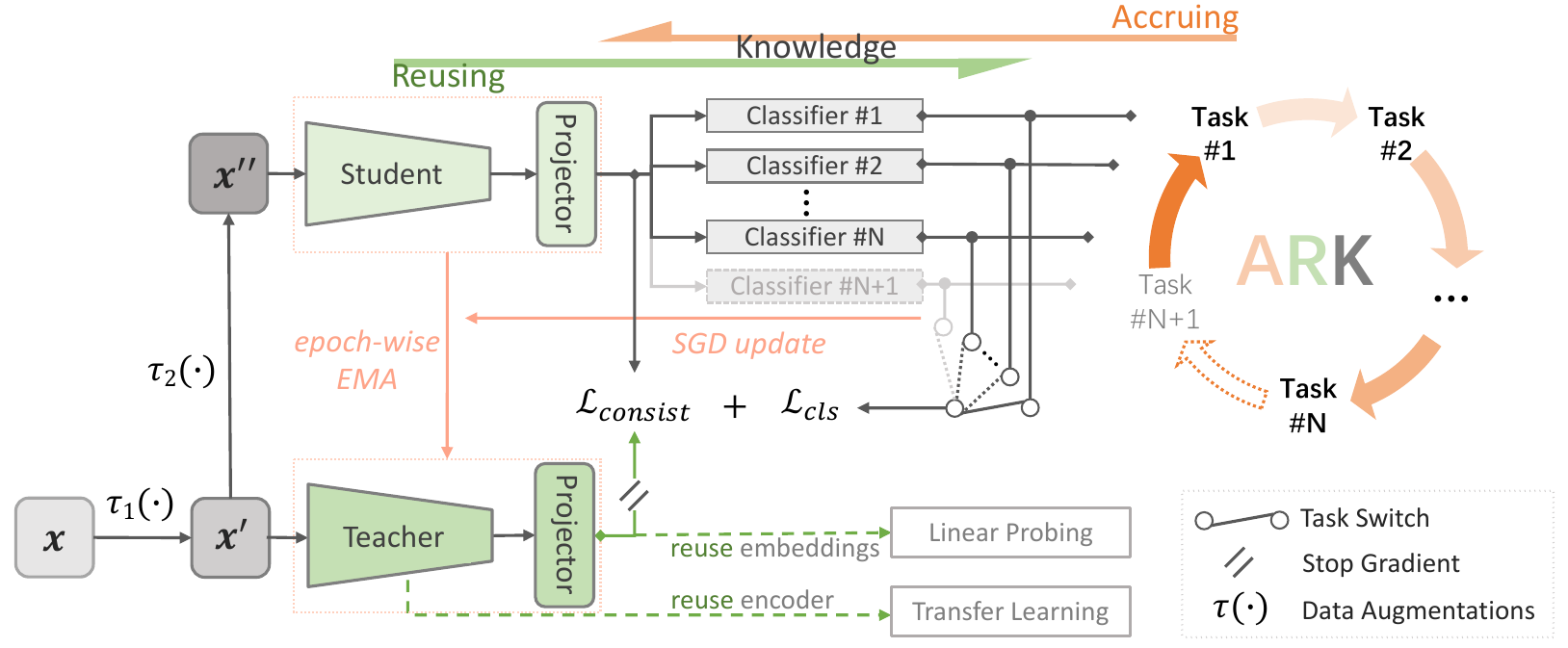}\centering
\caption{Our Ark is built on a student-teacher model with multi-task heads and trained via cyclic pretraining, aiming to accrue and reuse the expert knowledge embedded in the {\em heterogeneous} labels with numerous public datasets (see \sectionname~\ref{sec:ark} for details).}
\label{fig:ark}
\end{figure}

\smallskip
\noindent\textbf{Reusing accrued knowledge from the student to bolster cyclic pretraining. }
If the model learns from multiple tasks sequentially, it may ``forget'' the previously learned knowledge, and its performance on an old task may degrade catastrophically~\cite{kemker2018forgetting}. This problem is addressed naturally in Ark by cyclic pretraining, where the model revisits all the tasks in each round and reuses all knowledge accrued from the previous rounds and tasks to strengthen its learning from the current and future tasks. That is, by regularly reviewing the accrued knowledge through task revisitation, Ark not only prevents forgetting but also enables more efficient and effective learning from multiple tasks iteratively.

\smallskip
\noindent\textbf{Reusing accrued knowledge from the teacher to mitigate forgetting. }
To leverage the accumulated knowledge of the teacher model as an additional self-supervisory signal, we incorporate a consistency loss between the student and the teacher, as shown in~\figurename~\ref{fig:ark}. To enhance this supervision, we introduce projectors in Ark that map the outputs of the student and teacher encoders to the same feature space. This further reinforces the feedback loop between the student and teacher models, facilitating the transfer of historical knowledge from the teacher to the student as a reminder to mitigate forgetting.

\smallskip
\noindent{\ul{\textit{Ark has the following properties}}:}

\smallskip
\noindent\textbf{$\bullet$ Knowledge-centric.} Annotating medical images by radiologists for deep learning is a process of transferring their in-depth knowledge and expertise in interpreting medical images and identifying abnormalities to a medium that is accessible for computers to learn. Ark's superior and robust performance is attributed to the accumulation of expert knowledge conveyed through medical imaging annotations from diverse expert sources worldwide. At the core of Ark is acquiring and sharing knowledge: ``knowledge is power'' (Mac Flecknoe) and ``power comes not from knowledge kept but from knowledge shared'' (Bill Gates).

\smallskip
\noindent\textbf{$\bullet$ Label-agnostic, task-scalable and annotation-heterogeneous.} Ark is label-agnostic as it does not require prior label “understanding” of public datasets, but instead uses their originally-provided labels. It is designed with pluggable multi-task heads and cyclic pretraining to offer flexibility and scalability for adding new tasks without manually consolidating heterogeneous labels or training task-specific controllers/adapters~\cite{zhuassembling}. Therefore, Ark intrinsically handles the annotation heterogeneity across different datasets.

\smallskip
\noindent\textbf{$\bullet$ Application-versatile.} Ark trains versatile foundation models by utilizing a large number of publicly-available images from diverse sources and their readily-accessible diagnostic labels. As shown in \sectionname~\ref{sec:results}, Ark models are more robust, generalizable, and transferable to a wide range of application-specific target tasks across diseases (\eg pneumothorax, tuberculosis, cardiomegaly) and anatomies (\eg lung, heart, rib), highlighting Ark’s versatility.

\section{Experiments and Results}\label{sec:results}

Our Ark-5 and Ark-6 take the base version of the Swin transformer (Swin-B)~\cite{liu2021swin} as the backbone, feature five and six independent heads based on the pretraining tasks and their classes, and are pretrained on Datasets 1-5 and 1-6, respectively, with all validation and test data excluded to avoid test-image leaks. In the following, both models are evaluated via transfer learning (in~\sectionname~\ref{sec:results}.1 and \ref{sec:results}.2) on a wide range of 10 common, yet challenging, tasks on 8 publicly available datasets, encompassing various thoracic diseases and diverse anatomy. To provide a more comprehensive evaluation, we conduct linear probing (in~\sectionname~\ref{sec:results}.3) and analyze gender biases (in~\sectionname~\ref{sec:results}.4) on the Ark models in comparison with Google CXR-FM. 
Pretraining and evaluation protocols are detailed in Appendix~E.

\smallskip
\noindent\textbf{1) Ark outperforms SOTA fully/self-supervised methods on various tasks for thoracic disease classification.}

\noindent{\ul{\textit{Experimental Setup}}:}
To demonstrate the performance improvements achieved through Ark pretraining, we compare the Ark models with SOTA fully-supervised and self-supervised models~\cite{liu2021swin,xie2022simmim} that were pretrained on ImageNet. We also include a comparison with a SOTA domain-adapted model~\cite{Ma2022Benchmarking} that was first pretrained on ImageNet and then on a large-scale domain-specific dataset comprising 926,028 CXRs from 13 different sources.
All downstream models share the same Swin-B backbone, where the encoder is initialized using the pretrained weights and a task-specific classification head is re-initialized based on the number of classes for the target task. We fine-tune all layers in the downstream models under the same experimental setup. We also report the results of training the downstream models from scratch (random initialization) as the performance lower bound. Note that Google CXR-FM cannot be included for comparison as it is not publicly released for fine-tuning.

\noindent{\ul{\textit{Results and Analysis}}:}
As shown in~\tablename~\ref{tab:finetune}, our Ark models consistently outperform the SOTA fully/self-supervised ImageNet pretrained models on all target tasks. These results highlight the benefit of leveraging additional domain-relevant data in pretraining to reduce the domain gap and further improve the model's performance on target tasks. 
Furthermore, compared with the self-supervised domain-adapted model that utilizes 926K CXRs for pretraining, Ark models yield significantly superior performance on Dataset 1, 3-5 with only 335K CXRs, and on-par performance on 2.NIHC with 704K CXRs. These results demonstrate the superiority of Ark that accrues and reuses the knowledge retained in heterogeneous expert annotations from multiple datasets, emphasizing the importance of learning from expert labels. Moreover, we observe that Ark-6 consistently outperforms Ark-5, indicating the importance of incorporating more data and annotations from diverse datasets in pretraining.

\begin{table}[t]
\centering
\caption{Our Ark-5 and Ark-6 outperform SOTA ImageNet pretrained models and the self-supervised domain-adapted model that utilizes even more training data, highlighting the importance of accruing and reusing knowledge in expert labels from diverse datasets for both classification and segmentation. With the best bolded and the second best underlined, a statistical analysis is conducted between the best vs. others, where green-highlighted boxes indicate no statistically significant difference at level $p = 0.05$.}
\label{tab:finetune}
\scriptsize
\begin{tabular}{c P{0.143\linewidth}P{0.134\linewidth}P{0.134\linewidth}P{0.134\linewidth}P{0.134\linewidth}P{0.134\linewidth}}\toprule
\multirow{2}{*}{Initialization}&\multirow{2}{*}{Pretraining}  & \multicolumn{5}{c}{Classification task}\\
  & &1.CXPT &2.NIHC &3.RSNA & 4.VINC & 5.NIHS \\\midrule
Random &- &83.39\tiny±0.84 &77.04\tiny±0.34 &70.02\tiny±0.42 &78.49\tiny±1.00 &92.52\tiny±4.98 \\
Supervised &\tiny IN &87.80\tiny±0.42 &81.73\tiny±0.14 &73.44\tiny±0.46 &90.35\tiny±0.31 &93.35\tiny±0.77 \\
SimMIM &\tiny IN &88.16\tiny±0.31 &81.95\tiny±0.15 &73.66\tiny±0.34 &90.24\tiny±0.35 &94.12\tiny±0.96 \\
SimMIM &\tiny IN$\rightarrow$CXR(926K) &88.37\tiny±0.40 &\cellcolor{maroon!15}\underline{83.04\tiny±0.15} &74.09\tiny±0.39 &91.71\tiny±1.04 &95.76\tiny±1.79 \\
Ark-5\tiny(ours) &\tiny IN$\rightarrow$CXR(335K) &\underline{88.73\tiny±0.20} &82.87\tiny±0.13 &\cellcolor{maroon!15}\underline{74.73\tiny\tiny±0.59} &\underline{94.67\tiny\tiny±0.33} &\cellcolor{maroon!15}\underline{98.92\tiny\tiny±0.21} \\
Ark-6\tiny(ours) &\tiny IN$\rightarrow$CXR(704K) &\cellcolor{maroon!15}\textbf{89.14\tiny\tiny±0.22} &\cellcolor{maroon!15}\textbf{83.05\tiny\tiny±0.09} &\cellcolor{maroon!15}\textbf{74.76\tiny±0.35} &\cellcolor{maroon!15}\textbf{95.07\tiny±0.16} &\cellcolor{maroon!15}\textbf{98.99\tiny±0.16} \\
\bottomrule
\end{tabular}
\scriptsize
\begin{tabular}{c P{0.143\linewidth}P{0.134\linewidth}P{0.134\linewidth}P{0.134\linewidth}P{0.134\linewidth}P{0.134\linewidth}}\toprule
\multirow{2}{*}{Initialization}& \multirow{2}{*}{Pretraining}& \multicolumn{5}{c}{Segmentation task}\\
 & &7.NIHM &8.JSRT\tiny{Lung} &8.JSRT\tiny{Heart} & 8.JSRT\tiny{Clavicle} & 9.VINR \\\midrule
 
Random &- &96.32\tiny±0.18 &96.32\tiny±0.10 &92.35\tiny±0.20 &85.56\tiny±0.71 &56.46\tiny±0.62 \\
Supervised &\tiny IN &97.23\tiny±0.09 &97.13\tiny±0.07 &92.58\tiny±0.29 &86.94\tiny±0.69 &62.40\tiny±0.80 \\
SimMIM &\tiny IN &97.12\tiny±0.14 &96.90\tiny±0.08 &93.53\tiny±0.11 &87.18\tiny±0.63 &61.64\tiny±0.69 \\
SimMIM &\tiny IN$\rightarrow$CXR(926K) &97.10\tiny±0.40 &96.93\tiny±0.12 &93.75\tiny±0.36 &88.87\tiny±1.06 &\cellcolor{maroon!15}63.46\tiny±0.89 \\
Ark-5\tiny(ours) &\tiny IN$\rightarrow$CXR(335K) &\cellcolor{maroon!15}\underline{97.65\tiny±0.17} &\underline{97.41\tiny±0.04} &\underline{94.16\tiny±0.66} &\cellcolor{maroon!15}\underline{90.01\tiny±0.35} &\cellcolor{maroon!15}\textbf{63.96\tiny±0.30} \\
Ark-6\tiny(ours) &\tiny IN$\rightarrow$CXR(704K) &\cellcolor{maroon!15}\textbf{97.68\tiny±0.03} &\cellcolor{maroon!15}\textbf{97.48\tiny±0.08} &\cellcolor{maroon!15}\textbf{94.62\tiny±0.16} &\cellcolor{maroon!15}\textbf{90.05\tiny±0.15} &\cellcolor{maroon!15}\underline{63.70\tiny±0.23} \\
\bottomrule
\end{tabular}
\end{table}

\smallskip
\noindent\textbf{2) Ark provides generalizable representations for segmentation tasks.}

\noindent{\ul{\textit{Experimental Setup}}:}
To evaluate the generalizability of Ark's representations, we transfer the Ark models to five segmentation tasks involving lungs, heart, clavicles, and ribs, and compare their performance with three SOTA fully/self-supervised models.
We build the segmentation network upon UperNet~\cite{xiao2018upernet}, which consists of a backbone network, a feature pyramid network, and a decoder network. We implement the backbone network with Swin-B and initialize it with the pretrained weights from the Ark and those aforementioned SOTA models. The remaining networks are randomly initialized. We then fine-tune all layers in the segmentation models under the same experimental setup. 

\noindent{\ul{\textit{Results and Analysis}}:} 
As seen in~\tablename~\ref{tab:finetune}, Ark models achieve significantly better performance than the SOTA models, demonstrating that Ark learned generalizable representations for delineating organs and bones in CXR. This superior performance is achieved by pretraining using large-scale CXRs and various disease labels from diverse datasets. Clinically, certain thoracic abnormalities can be diagnosed by examining the edges of the lungs, heart, clavicles, or ribs in CXR. For instance, a pneumothorax can be detected by observing a visible ``visceral pleural line'' along part or all of the length of the lateral chest wall~\cite{mason2010murray}. Cardiomegaly can be diagnosed when the heart appears enlarged, with maximum diameter of the heart exceeding a pre-defined cardiothoracic ratio~\cite{wang2017nihchestx}. Fractures can be identified when the edges of the clavicles or ribs appear abnormally displaced or the bone cortex appears offset~\cite{collins2000chest}. Therefore, leveraging diagnostic information from disease labels during pretraining enables Ark models to better capture the nuanced and varied pathological patterns, strengthening the models' ability to represent anatomically specific features that reflect abnormal conditions in various oragns or bones. By contrast, the SimMIM (IN$\rightarrow$CXR(926K)) model is pretrained with a self-supervised masked image modeling proxy task, which may use many clues to reconstruct the masked patches that are not necessarily related to pathological conditions, leading to lower performance despite training on more images.

\smallskip
\noindent\textbf{3) Ark offers embeddings with superior quality over Google CXR-FM.} 

\noindent{\ul{\textit{Experimental Setup}}:}
To highlight the benefits of learning from more detailed diagnostic disease labels, we compare our Ark models with Google CXR-FM. CXR-FM was trained on a large dataset of 821,544 CXRs from three different sources, but with coarsened labels (normal or abnormal). By contrast, our Ark models are trained with less data, but aims to fully utilize all labels provided by experts in the original datasets. Furthermore, Ark models employ a much smaller backbone (88M parameters) compared with CXR-FM using EfficientNet-L2 (480M parameters). 
Since Google CXR-FM is not released and cannot be fine-tuned, we resorted to its released API to generate the embeddings (information-rich numerical vectors) for all images in the target tasks. For the sake of fairness, we also generated the embeddings from Ark's projector, whose dimension is the same as Google's. 
To evaluate the quality of the learned representations of these models, we conduct linear probing by training a simple linear classifier for each target task. The performance of both models is evaluated on six target tasks, including an unseen dataset, 10.SIIM, where the images have not been previously seen by the Ark models during pretraining. Additionally, we perform the same evaluation on 10.SIIM with partial training sets or even few-shot samples to further demonstrate the high quality of our Ark models' embeddings.

\begin{figure}[t]
\includegraphics[width=\textwidth]{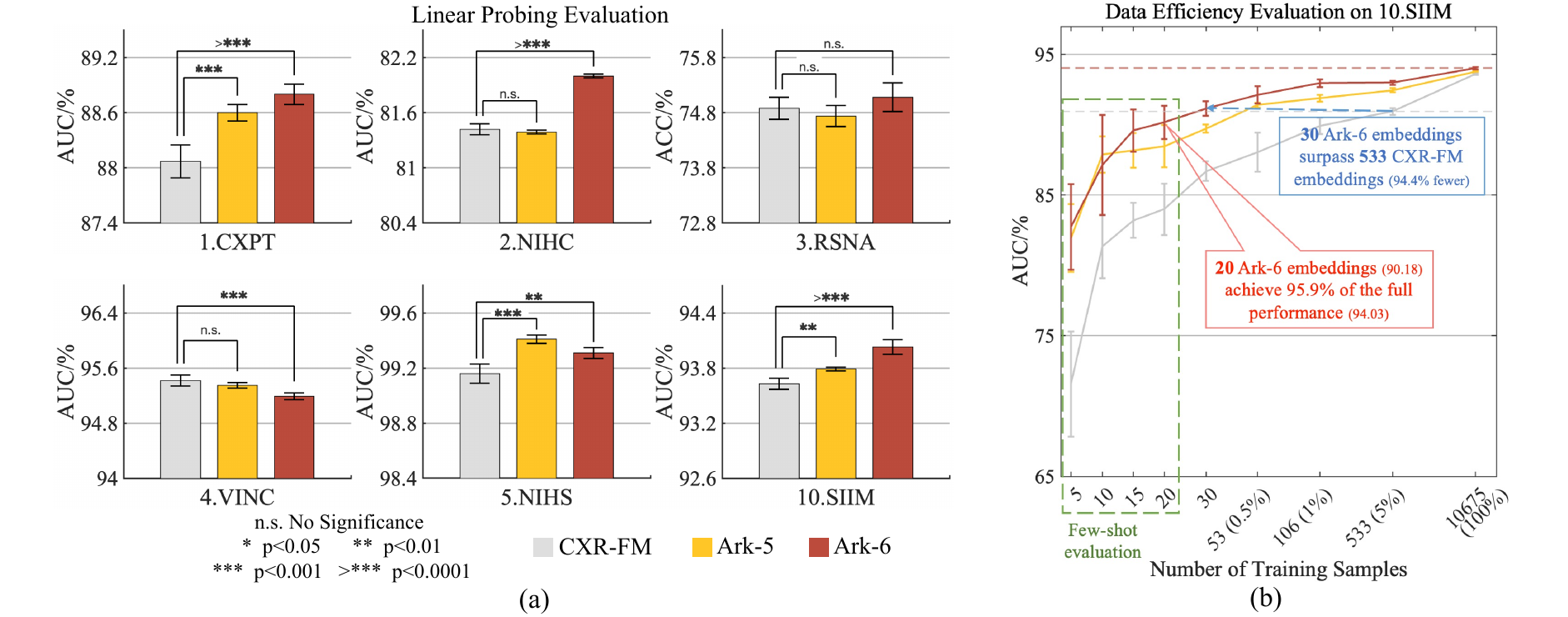}\centering
\caption{Ark-5 and Ark-6 are compared with Google CXR-FM via linear probing (a) with complete training set on six target tasks, demonstrating Ark's superior or comparable performance and better embedding quality, and (b) with partial training sets or even few-shot samples, showcasing Ark's outstanding performance in terms of data efficiency.} 
\label{fig:linear}
\end{figure}

\noindent{\ul{\textit{Results and Analysis}}:} 
\figurename~\ref{fig:linear}(a) shows that Ark-6 outperforms CXR-FM significantly on Dataset 1,2,5 and 10, and performs comparably to CXR-FM on 3.RSNA. Similarly, Ark-5 performs better than CXR-FM on Dataset 1,5 and 10, while performing comparably on the remaining tasks. Moreover,~\figurename~\ref{fig:linear}(b) shows that both Ark-5 and Ark-6 consistently outperform CXR-FM in small data regimes, highlighting the superiority of Ark's embeddings, which carry richer information that can be utilized more efficiently. These results demonstrate that Ark models learn higher-quality representations with less pretraining data while employing a much smaller backbone than CXR-FM, highlighting that learning from more granular diagnostic labels, such as Ark, is superior to learning from coarsened normal/abnormal labels.

\smallskip
\noindent\textbf{4) Ark shows a lower false-negative rate and less gender bias. }

\noindent{\ul{\textit{Experimental Setup}}:}
Underdiagnosis, as indicated by false-negative results, may lead to delayed treatment, resulting in severe consequences. Moreover, population-imbalanced data may train models of biases, adversely affecting diagnostic performance in minority populations. Therefore, a robust computer-aided diagnosis (CAD) system should provide a low false-negative rate and strong resilience to biased training data. To demonstrate our Ark’s robustness relative to Google CXR-FM, we first examine their gender-based FNRs on 1.CXPT and 2.NIHC, as these two datasets offer sufficient cases with gender information. We further investigate gender biases in Ark-6 and CXR-FM on 1.CXPT using gender-exclusive training sets by following the setup in~\cite{larrazabal2020gender}. We use their training/test splits to ensure a balanced number of cases per class in 40 male/female-only folds, train linear classifiers on those folds using embeddings from Ark-6 and CXR-FM, and then evaluate these classifiers on the corresponding male/female-only test splits. A model, if biased, will perform differently when its training and test data are of the opposite gender. We detail this setup in Appendix~E.

\begin{figure}[t]
\includegraphics[width=\textwidth]{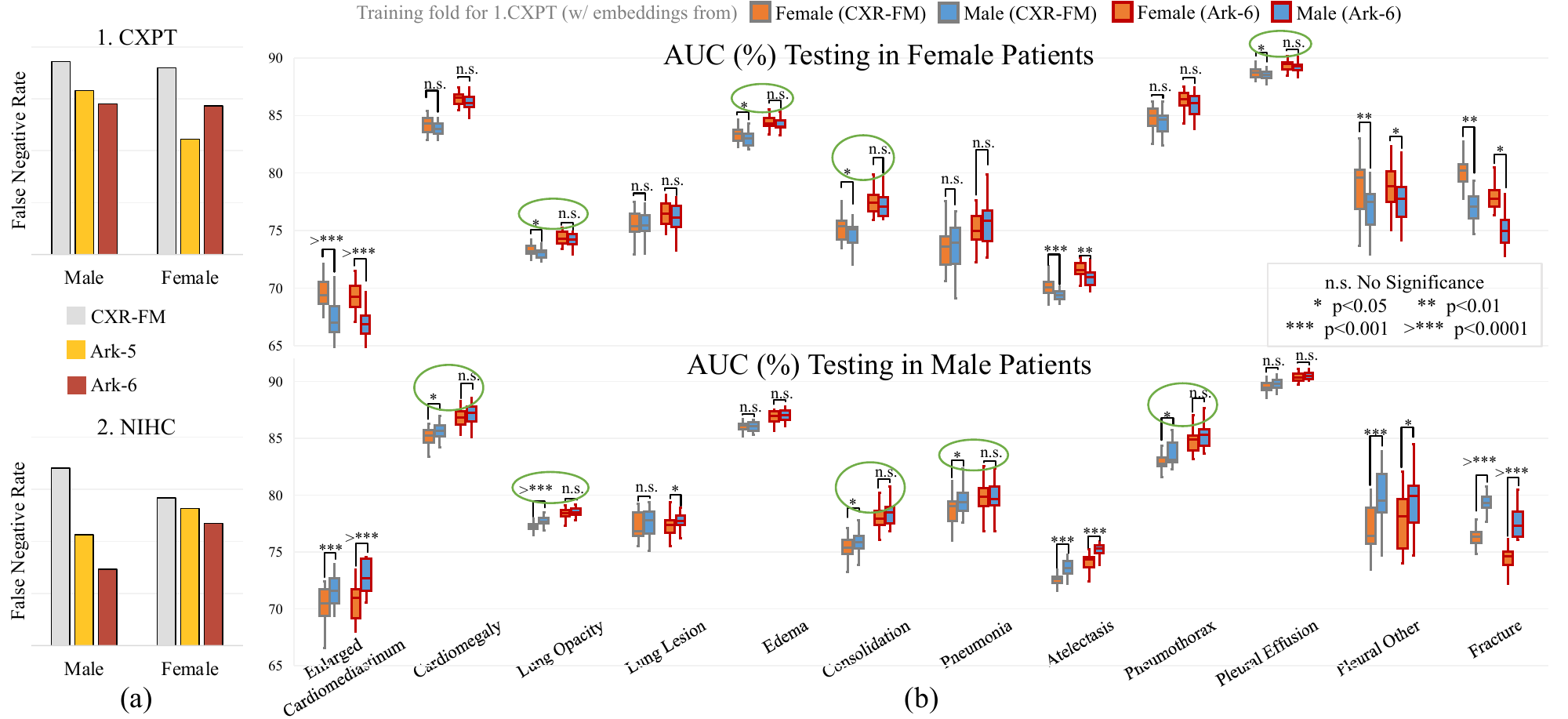}\centering
\caption{Ark models are compared with Google CXR-FM as regards false-negative rate (FNR) and gender-related bias. (a) Ark models show lower FNRs, indicating superior underdiagnosis mitigation. (b) Ark-6 demonstrates greater resilience to gender-imbalanced data. Gender bias is characterized by a significant drop in performance when training and test data are of the opposite gender, compared to when they are of the same gender (\eg the orange whisker boxes are lower than the blue boxes in the lower-part (b)). Each green circle indicates a lung disease with gender bias by CXR-FM, as it performs differently between training on male and female data. But Ark exhibits a more robust performance, showing no significant difference on gender-segregated data.} 
\label{fig:bias}
\end{figure}

\noindent{\ul{\textit{Results and Analysis}}:} 
\figurename~\ref{fig:bias}(a) illustrates that Ark models have lower FNRs than CXR-FM for both genders on both tasks, demonstrating that Ark models are less likely to underdiagnose disease conditions than CXR-FM. 
In~\figurename~\ref{fig:bias}(b), the biases in the pretrained models are measured by performance differences between linear classifiers trained on male-only and female-only embeddings. The upper part of~\figurename~\ref{fig:bias}(b) depicts the results of {\em testing on female-only} sets, where the classifiers {\em trained on male-only }embeddings generally perform poorly compared with those trained on female embeddings, revealing gender biases due to data imbalance. Among the 12 diseases, the classifiers trained with Google's embeddings have unbiased performances for only 4 diseases, whereas those using Ark-6's embeddings perform in an unbiased fashion with no significant differences for the 8 diseases. 
The same situation occurs when testing is performed on male patients as shown in the lower part of~\figurename~\ref{fig:bias}(b). The gender bias analysis demonstrates that Ark has greater robustness to the extremely imbalanced data that contributes to gender bias in computer-aided diagnosis.

\section{Conclusions and Future Work}

We have developed Foundation Ark, the first open foundation model, that realizes our vision: accruing and reusing knowledge retained in heterogeneous expert annotations with numerous datasets offers superior and robust performance. Our experimental results are strong on CXRs, and we plan to extend Ark to other modalities. We hope Ark's performance encourages researchers worldwide to share codes and datasets big or small for creating open foundation models, accelerating open science, and democratizing deep learning for medical imaging.


\bibliographystyle{splncs04}
\bibliography{references}

\include{SupplementaryMaterial}

\end{document}

%% file: SupplementaryMaterial.tex
\newpage

\title{Supplementary Material\\Foundation Ark: Accruing and Reusing Knowledge for Superior and Robust Performance}
\titlerunning{Foundation Ark: Accruing and Reusing Knowledge}
\author{DongAo Ma\inst{1} \and 
 Jiaxuan Pang\inst{1} \and
Michael B. Gotway\inst{2} \and
Jianming Liang\inst{1}} 
\authorrunning{D. Ma et al. }
\institute{Arizona State University, Tempe, AZ 85281, USA 
\email{\{dongaoma,jpang12,jianming.liang\}@asu.edu} \and
 Mayo Clinic, Scottsdale, AZ 85259, USA\\
\email{Gotway.Michael@mayo.edu}}
\maketitle
\appendix
\vspace{-2 em}
\begin{abstract}
    This supplementary material complements the paper titled ``Foundation Ark: Accruing and Reusing Knowledge for Superior and Robust Performance''. It is organized as follows. In \sectionname~\ref{appendix:A}, we present a comprehensive list of diagnostic labels from different public datasets, revealing marked label heterogeneity across institutions. \sectionname~\ref{appendix:B} offers a comparison between Ark and other existing works concerning the assembly of public datasets, emphasizing Ark's label-agnostic and task-scalable advantages. Section \ref{appendix:C} includes ablation studies that demonstrate the necessity of the projector and consistency loss, along with the superiority of the teacher model. In Sections \ref{appendix:D} and \ref{appendix:E}, we present pseudocode for Ark's cyclic pretraining and elaborate on the experimental setups. Lastly, Section \ref{appendix:F} contains acknowledgments for support.
\end{abstract}
\vspace{-2.2 em}
\begin{eqnarray*}
    \mbox{\scriptsize \bf Knowledge is power}&\mbox{---}&\mbox{\scriptsize Mac Flecknoe} \\[-4pt]
    \mbox{\scriptsize \bf Power comes not from knowledge kept but from knowledge shared}&\mbox{---}&\mbox{\scriptsize Bill Gates}
\end{eqnarray*}    

\vspace{-1.5 em}
\section{Heterogeneous Labels}\label{appendix:A}
\vspace{-2 em}
\begin{table}[!htb]\centering
\renewcommand\thetable{3}
\caption{As listed in this table, datasets created at different institutions tend to be annotated differently even when addressing the same clinical issue. Our Ark aims to accrue and reuse expert knowledge from {\em heterogeneous} labels with numerous public datasets to pretrain generic source models that are more robust, generalizable, and transferable to application-specific target tasks, demonstrating superior and robust performance over the SOTA fully/self-supervised baselines (\tablename~\ref{tab:finetune}) and Google CXR-FM (\figurename~\ref{fig:linear}). The challenge of learning from heterogeneous labels is addressed in Ark via multi-task heads and cyclic pretraining (\figurename~\ref{fig:ark}).} \label{tab:labels}
\vspace{-0.5 em}  
\scriptsize
\begin{tabular}{l P{0.873\linewidth}}\toprule
 Dataset  & Inconsistencies in diagnostic labels associated with popular public X-rays datasets\\\midrule
\multirow{3}{*}{\makecell[l]{1.CXPT\\6.MMIC}} & \multirow{3}{*}{\makecell[c]{No Finding, Enlarged Cardiomediastinum, Cardiomegaly, Lung Opacity, Lung Lesion,\\ Edema, Consolidation, Pneumonia, Atelectasis, Pneumothorax, Pleural Effusion, \\Pleural Other, Fracture, Support Devices}}\\
&\\&\\
\multirow{3}{*}{2.NIHC} &\multirow{3}{*}{\makecell[c]{Atelectasis, Cardiomegaly, Effusion, Infiltration, Mass, Nodule, Pneumonia, Pneumo-\\thorax, Consolidation, Edema, Emphysema, Fibrosis, Pleural Thickening, Hernia}} \\
&\\ & \\
3.RSNA  &Normal, No Lung Opacity/Not Normal, Lung Opacity \\
\multirow{2}{*}{4.VINC}  &\multirow{2}{*}{\makecell[c]{Pleural Effusion, Lung Tumor, Pneumonia, Tuberculosis, Other Diseases, No Finding}} \\
&\\
5.NIHS  &Tuberculosis \\
\bottomrule
\end{tabular}
\end{table}

\newpage
\section{Other Works for Assembling Public Datasets}\label{appendix:B}
\begin{table}[!htb]\centering
\renewcommand\thetable{4}
\caption{Our Ark is \ul{dataset/task-agnostic} as it does not require prior label ``understanding'' of public datasets. Unlike the listed example works that need to manually assemble the labels into a pre-defined list and train a dynamic controller/adapter as directives for different tasks, Ark is designed with pluggable multi-task heads and cyclic pretraining (\sectionname~\ref{sec:ark}) to offer flexibility and scalability for adding new tasks without manually consolidating heterogeneous labels or training task-specific controllers/adapters.}
\label{tab: related}

\scriptsize
\begin{tabular}{cP{0.35\linewidth}P{0.45\linewidth}}\toprule
Related works & \makecell[c]{How to preprocess labels?}  & \makecell[c]{When a new task comes?} \\\midrule
\multirow{2}{*}{\makecell[c]{Label-Assemble\tablefootnote[1]{\scriptsize{Zhu \etal (2022), Assembling Existing Labels from Public Datasets to Diagnose Novel Diseases: COVID-19 in Late 2019}}}} & 
\multirow{2}{*}{\makecell[c]{Need a pre-defined label list}}  & 
Update the label list and retrain the adapter if any labels aren't in the original list \\
&&\\

\multirow{2}{*}{\makecell[c]{DoDNet\tablefootnote[2]{\scriptsize{Zhang \etal (2020), DoDNet: Learning to segment multi-organ and tumors from multiple partially labeled datasets}}}} & 
\multirow{2}{*}{\makecell[c]{Need a pre-defined task list}} &
Renew the task list and retrain the controller when adding new tasks\\
&&\\

\multirow{2}{*}{\makecell[c]{CLIP-diven\tablefootnote[3]{\scriptsize{Liu \etal (2023), CLIP-Driven Universal Model for Organ Segmentation and Tumor Detection}}}} &
Need manual designs of prompt to get CLIP embeddings & 
Re-generate the CLIP embedding for any new classes and retrain the controller\\
&&\\

\multirow{3}{*}{Ark}  & 
\multirow{3}{*}{\makecell[c]{Task-agnostic, use all\\ readily-accessible labels directly \\ as they are}} &
Plug in Ark a new head, independent from existing tasks, for the new task, no modification on the rest architecture \\

\bottomrule
\end{tabular}
\end{table}

\section{Ablation study}\label{appendix:C}
\begin{table}[!htb]\centering
\renewcommand\thetable{5}
\caption{Our ablation studies on Ark-5 via linear probing show the projector and consistency loss are essential and the teacher significantly outperforms the student.} \label{tab:ablation}  

\scriptsize
\begin{tabular}{P{0.134\linewidth}P{0.134\linewidth}P{0.134\linewidth}P{0.134\linewidth}P{0.134\linewidth}P{0.134\linewidth}P{0.134\linewidth}}\toprule
Model &Projector & $\mathcal{L}_{consist}$ &2.NIHC &3.RSNA &4.VINC &5.NIHS \\\midrule
Teacher &$\times$ &$\times$ &81.09\tiny±0.08 &74.21\tiny±0.42 &94.89\tiny±0.07 &98.81\tiny±0.25 \\
Teacher &$\times$ &\checkmark &81.19\tiny±0.05 &74.42\tiny±0.25 &95.24\tiny±0.08 &99.01\tiny±0.08 \\
Student &\checkmark &\checkmark &81.34\tiny±0.04 &74.12\tiny±0.11 &94.85\tiny±0.07 &99.17\tiny±0.07 \\
Teacher &\checkmark &\checkmark &\textbf{81.39\tiny±0.02} &\textbf{74.74\tiny±0.19} &\textbf{95.35\tiny±0.04} &\textbf{99.41\tiny±0.03} \\
\bottomrule
\end{tabular}
\end{table}

\newpage
\section{Pseudocode for Ark's cyclic pretraining}\label{appendix:D}
As illustrated in \figurename~\ref{fig:ark} and described in Algorithm~\ref{alg:ark}, Ark is built on a teacher-student model, whose student is augmented with multi-task heads (each corresponding to one task) and trained via cyclic pretraining. Cyclic pretraining is an iterative process. At each iteration, the student aims to accrue knowledge from every expert annotation through its corresponding task head by sequentially scanning all datasets (tasks) one by one for one epoch. At the end of each task, the accrued knowledge is accumulated into the teacher (via EMA) and reused to help accrue more knowledge from the expert annotations associated with the next dataset. To reinforce the feedback loop between the student and teacher, after their encoders, a projector is introduced to map the representations to the same feature space via the consistency loss, also serving as the embedding for linear probing in our evaluation. After pretraining, the accumulated knowledge in the teacher is reused and transferred to the application-specific target tasks. 

\begin{algorithm}[htb]
\DontPrintSemicolon
\KwData{Datasets: $\mathcal{D}=\{D_1, D_2, ..., D_n\}$; Sample: image-label pair $(x,y) \in \mathcal{D}_i$}
\KwFunctions{Data augmentation: $\tau_1(\cdot)$, $\tau_2(\cdot)$; Dataset/task-specific losses: $\{\mathcal{L}_{D1}(\cdot,\cdot), \mathcal{L}_{D2}(\cdot,\cdot), ..., \mathcal{L}_{Dn}(\cdot,\cdot)\}$; Consistency loss: $\mathcal{L}_{const}(\cdot,\cdot)$; Loss update by SGD optimizer: $Update_{sgd}(\cdot,\cdot)$ }
\KwTrainables{Student's encoder and projector: $e_s$, $p_s$;  Multi-task heads $\mathcal{H} = \{h_1, h_2, ..., h_n\}$ }
\KwStopgrad{Teacher's encoder and projector: $e_t$, $p_t$}
\KwHyperparameters{Momentum: $\lambda$}

 $\{e_t, p_t\} \leftarrow \{e_s, p_s\}$ \tcp*{initialize teacher with student's parameters}
\For{$D_i$ \textbf{in} ${D_1, D_2, ..., D_n}$}   
{ 
    \tcc{train student for one epoch}
    \For{$(x,y)$ \textbf{in} $D_i$} 
    {
        $x' = \tau_1(x)$\;
        $x'' = \tau_2(x')$\;
        $emb_t,  emb_s = p_t(e_t(x')),  p_s(e_s(x''))$\;
        $pred = h_i(emb_s)$\;
        $Loss = \mathcal{L}_{Di}(pred, y) + \mathcal{L}_{const}(emb_t,emb_s)$ \;
        $Update(\{e_s, p_s, h_i\}, Loss) $\;
    }
    \tcc{Update teacher by student's parameters via epoch-wise EMA}
    $\{e_t, p_t\} \leftarrow \lambda  \{e_t, p_t\} + (1-\lambda) \{e_s, p_s\}$
}

\caption{A round of Ark's cyclic pretraining}\label{alg:ark}
\end{algorithm}

\section{Experiment details}\label{appendix:E}

\noindent\textbf{Pretraining:}
We have trained Ark-5/6 with 335,484/704,363 chest X-rays from the first 5/6 datasets in \tablename~\ref{tab:datasets} collected by 5/6 different institutions around the world and annotated by their experts. We use their originally-provided labels (\tablename~\ref{tab:labels}), showing marked differences across institutions. To avoid test-image leaks, all
validation and test data are excluded from the Ark pretraining. We employ the base version of the Swin transformer with an input resolution of $224\times224$ as the backbone. The encoders in teacher and student are initialized with the officially released weights trained on ImageNet\footnote[1]{\scriptsize{\url{GitHub.com/SwinTransformer/storage/releases/download/v1.0.0/swin_base_patch4_window7_224_22kto1k.pth}}}, and the projectors and the multi-task heads are randomly initialized. 
The task-specific (classification) loss is associate with each dataset based on its labels. We use binary cross-entropy for the binary/multi-label classification tasks (Dataset 1-2, 4-6) and cross-entropy for the multi-class classification task (Dataset 3).
Besides, we use mean-squared error for the consistency loss. 
We optimize the student model using SGD optimizer with an initial learning rate of 0.3, and a batch size of 200 distributed across 4 Nvidia V100 GPUs with a memory of 32 GB per-card; we apply a \textit{stop-gradient} operator on the teacher and update it using \textit{epoch-wise EMA} of the student parameters at the end of each task with an initial momentum of 0.9. The image augmentation function $\tau_1(.)$ includes random cropping and rotation, and $\tau_2(.)$ includes randomly changing brightness, contrast, and Gamma distribution of an image.

\smallskip\noindent\textbf{Evaluation:}
We have evaluated Ark-5 and Ark-6 via transfer learning and compared them with SOTA fully-supervised and self-supervised models (\tablename~\ref{tab:finetune}). For fair comparisons, we follow the SoTA\footnote[2]{\scriptsize{\url{GitHub.com/JLiangLab/BenchmarkTransformers}}} and apply the same augmentations for all methods. We measure the performance of binary/multi-label classification by AUC (area under the ROC curve), multi-class classification by accuracy, and segmentation by Dice. We perform at least 10 trials, report the mean and standard deviation of the performance metrics, and further present statistical analysis based on an independent two-sample \textit{t}-test.

To provide a more comprehensive evaluation, we have conducted linear probing (\figurename~\ref{fig:linear}) and analyzed gender biases (\figurename~\ref{fig:bias}) on the Ark models in comparison with Google CXR-FM. We pre-generate the embeddings for all images in the target tasks from Ark-5, Ark-6 and Google CXR-FM\footnote[3]{\scriptsize{\url{GitHub.com/Google-Health/imaging-research/tree/master/cxr-foundation}}}, and then train a simple linear classifier for each target task. 

The evaluation of gender bias robustness for each model follows the above-mentioned linear probing protocol. We compute the FNRs in terms of gender using the linear probing results on 1.CXPT and 2.NIHC. Note that we could not analyze gender biases on datasets 3, 4, and 10 because they don’t come with patient genders. 
Furthermore, to analyze the gender biases in our Ark model and CXR-FM, we follow the train/test splits in the GenderBias\_CheXNet repository\footnote[4]{\scriptsize{\url{GitHub.com/N-Nieto/GenderBias\_CheXNet}}} to ensure a balanced number of cases per class in the 20 male-only and 20 female-only folds, where the labels ``No Finding'' and ``Support Device'' are excluded. We train 40 linear classifiers on the male-only and female-only splits using embeddings from Ark-6 and CXR-FM to evaluate their gender biases. We then evaluate these classifiers on the corresponding male/female-only test splits and report the average performance over the 20 folds.

\medskip\noindent\tablename~\ref{tab:setup} lists the key setups used in Ark's pretraining and evaluation protocols.

\begin{table}[t]\centering
\renewcommand\thetable{5}
\caption{ Experimental configuration details.}\label{tab:setup} 
\scriptsize
\begin{tabular}{lll}\toprule
\multicolumn{2}{l}{ \textbf{Ark pretraining setup}} \\\midrule
Backbone &Swin Transformer Base (input resolution: 224 $\times$ 224) \\ &\\
\multirow{2}{*}{Initialization} &Encoders: officially released ImageNet weights\\
&Projectors and Multi-task heads: random weights \\ &\\
\multirow{3}{*}{Loss function} & Task-specific loss: binary cross-entropy (BCE) for Dataset 1-2, 4-6 \\
                      & ~~~~~~~~~~~~~~~~~~~~~~~~cross-entropy (CE) for Dataset 3\\
&Consistency loss: mean-squared error (MSE) \\ &\\
\multirow{2}{*}{Optimization} &Student: SGD optimizer, learning rate of 0.3, Cosine scheduler \\
&Teacher: Stop gradient, EMA update, momentum of 0.9 \\ &\\
Pretraining & 200 rounds (iterates through all datasets 200 times)\\ &\\
\multirow{2}{*}{Augmentation} & $\tau_1(.)$: Random cropping and rotation \\
& $\tau_2(.)$: Random changing of image brightness, contrast, and Gamma distribution \\ &\\
Devices &4 Nvidia V100 GPUs (32GB) \\

\toprule
\multicolumn{2}{l}{ \textbf{Ark evaluation setup}} \\
\midrule
Tranferred model & Teacher's encoder\\ &\\
Embeddings & Pre-generated from Ark's projector with a dimension of $1\times1376$\\ &\\
\multirow{3}{*}{Metrics} & Binary/Multi-label classification:  Area under the ROC curve (AUC)\\
& Multi-class classification: Accuracy (ACC)\\
& Segmentation: Dice similarity coefficient (Dice)\\ & \\
Performance & Mean and Standard Deviation of the metrics for 10 trials\\ &\\
Significance test& Independent two-sample \textit{t}-test (p-value < 0.05)\\
\bottomrule
\end{tabular}
\end{table}

\section{Acknowledgements}\label{appendix:F}

This research has been supported in part by ASU and Mayo Clinic through a Seed Grant and an Innovation Grant, and in part by the NIH under Award Number R01HL128785. The content is solely the responsibility of the authors and does not necessarily represent the official views of the NIH. This work has utilized the GPUs provided in part by the ASU Research Computing and in part by the Bridges-2 at Pittsburgh Supercomputing Center through allocation BCS190015 and the Anvil at Purdue University through allocation MED220025 from the Advanced Cyberinfrastructure Coordination Ecosystem: Services \& Support (ACCESS) program, which is supported by National Science Foundation grants \#2138259, \#2138286, \#2138307, \#2137603, and \#2138296. We also acknowledge Google for granting us access to CXR Foundation API, which enabled us to generate the embeddings for the target datasets. The content of this paper is covered by patents pending.